\documentclass[letterpaper]{article} %DO NOT CHANGE THIS
\usepackage{ijcai19}  %Required
\usepackage{times}  %Required
\usepackage{helvet}  %Required
\usepackage{courier}  %Required
\usepackage{url}  %Required
\usepackage{graphicx}  %Required
\usepackage{color}
\usepackage{subfig}
\usepackage{amsmath}
\usepackage{amsfonts}
\usepackage{comment}
\usepackage{booktabs}
\usepackage{wrapfig}
\DeclareMathOperator*{\argmax}{argmax}

\title{Learning to Interpret Satellite Images in Global Scale Using Wikipedia}
\author{
Burak Uzkent$^1$\footnote{Contact Author},
Evan Sheehan$^1$,
Chenlin Meng$^1$,
Zhongyi Tang$^2$,
David Lobell$^2$,
Marshall Burke$^2$,
Stefano Ermon$^1$
\\
\affiliations
$^1$Department of Computer Science, Stanford University \\
$^2$Department of Earth Systems Science, Stanford University \\
\emails
buzkent@cs.stanford.edu, \{esheehan, chenlin, zztang, dlobell, mburke\}@stanford.edu, ermon@cs.stanford.edu
}

\begin{document}

\maketitle

\begin{abstract}
Despite recent progress in computer vision, fine-grained interpretation of satellite images remains challenging because of a lack of labeled training data. To overcome this limitation, 
%we propose using Wikipedia as a previously untapped source of rich, georeferenced textual information with global coverage. 
we construct a novel 
%global, multi-modal 
dataset called WikiSatNet by pairing geo-referenced Wikipedia articles with satellite imagery of their corresponding locations. We then propose two strategies to learn representations of satellite images by predicting properties of the corresponding articles from the images. Leveraging this new multi-modal dataset, 
%two different pre-training methods ; (1) weak supervision using handcrafted labels, and (2) unsupervised learning using image to text matching. 
%We then fine-tune the pre-trained models on a human-annotated dataset and demonstrate that our pre-training methods 
we can drastically reduce the quantity of human-annotated labels and time required for downstream tasks. 
%On the other hand, 
On the recently released fMoW dataset, 
%that contains about 350,000 human-annotated training samples, 
our pre-training strategies can boost the performance of a model pre-trained on ImageNet by up to $4.5\%$ in F1 score.
\end{abstract}

\section{Introduction}
%\input{intro2}
%\s{general note: pls comment out comments without deleting them so we can keep track of them}

Deep learning has been the driving force behind many recent improvements in computer vision tasks, including image classification, image segmentation, object detection and tracking, etc.~\cite{russakovsky2015imagenet,lin2014microsoft,han2018advanced,uzkent2017aerial,uzkent2016real,uzkent2013feature}. These deep models, however, require training on high quality, large-scale datasets, and building these datasets is typically very costly. 
%Deep Learning has also been effective in analyzing the satellite images. As in the other domains, we need large-scale datasets to train these models to analyze satellite images. 
%In particular, 
Satellite images are particularly difficult and expensive to label because of humans' unfamiliarity with aerial perspectives~\cite{christie2018functional}.

One effective way to reduce the amount of training data needed is to perform pre-training on an existing, previously annotated dataset, such as ImageNet~\cite{deng2009imagenet}, and transfer the learned weights to the domain of interest \cite{raina2007self,dai2009eigentransfer,uzkent2018tracking}. 
%\s{add some citations}
However, the success of this approach diminishes if the underlying distributions and/or compositions of the pre-training and target datasets are not sufficiently similar. Such a problem is exceptionally pronounced in the satellite imagery space, as the entire frame of reference and perspective of an aerial image is altered compared to a natural image.
%, making them completely dissonant. 
This has the unfortunate effect of rendering natural image datasets, such as ImageNet, less useful as pre-training mechanisms 
%for downstream, satellite-related, computer vision tasks 
for downstream computer vision tasks in the satellite domain~\cite{pan2010survey,kaiser2017learning}.

Because direct annotation is expensive, researchers have considered many creative ways to provide supervision without explicit labels. These include unsupervised~\cite{kingma2014semi}, %\s{cite some generative modeling stuff},
label-free 
\cite{DBLP:journals/corr/abs-1805-10561,stewart2017label}, and weakly supervised learning methods~\cite{ratner2017snorkel}. A particularly effective strategy is to leverage co-occurrence statistics in a dataset, e.g.,  predict the next frame in a video, a missing word in a sentence \cite{DBLP:journals/corr/abs-1301-3781},
%, encourage a model to learn similar representations in the nearby satellite images \cite{jean2018tile2vec}, %\neal{this wasn't done in our paper}
or predict relationships between entities such as images and text co-occurring together.
%utilize multi-modal datasets to explore embedded relations between entities (e.g., co-occurrences between images and text). 
%
%Another interesting study to pre-train a model using weakly supervised is proposed by \cite{mahajan2018exploring}.
%
%\neal{This sentence sounds a bit weird - it starts by saying ``a particularly effective strategy'', then lists some strategies that sound quite different, the last of which seems to lead into the next sentence.}
%\s{because direct annotation is expensive, researchers have considered many creative way to provide supervision without explicit labels.}
%\s{these include unsupervised learning, weakly supervised learning [cite chris re stuff, our stuff on constraints, etc], etc.}
%\s{a particularly effective way is to leverage co-occurrence statistics. e.g., predict next frame in a video, a missing word from a sentence [word2vec], a missing tile from an image [tile2vec]}.
%To reduce these expensive barriers, we can rely on other open source datasets to label the samples in the pre-training dataset. 
%\s{also Multi-modal datasets: e.g., co-occurrence between images and text}
%\burak{this two sentences does not connect well.}
For example, leveraging images and their hashtags on Instagram, \cite{mahajan2018exploring} build a large scale image recognition dataset consisting of more than 3 billion images across 17,000 weak labels obtained from textual hashtags and their WordNet \cite{miller1995wordnet} synsets. After pre-training on this extremely large dataset, they report almost $5\%$ improvement over the same model trained from scratch on ImageNet. 
%Unsupervised and weakly supervised learning on satellite images, on the other hand, have not been studied extensively. 

%\cite{jean2018tile2vec} use the triplet loss to pre-train a CNN on aerial images. Their approach assumes that spatially nearby images should contain similar visual structures. Applyinging this hyphothesis, they sample triplets from spatially nearby and distant regions and report improved performance when fine-tuning on the target task with perfect labels.

%\s{shorten this to a sentence or two}

%In addition to the natural images, deep learning has also enjoyed high popularity across the satellite imagery space in the domains of object detection, image recognition, etc., with the availability of large-scale datasets such as xView \cite{lam2018xview}, and fMoW \cite{christie2018functional}. When considering an object detection task, one could attempt to pre-train a deep learning model on the ImageNet dataset and fine-tune the weights on the xView dataset. 

\begin{figure*}[h]
\centering
\includegraphics[width=0.95\textwidth]{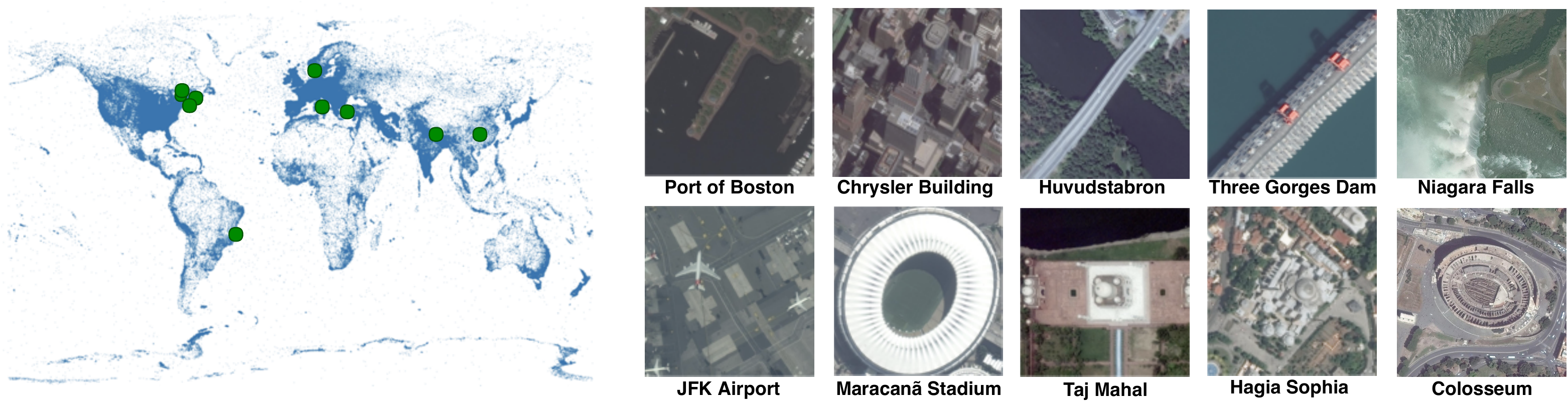}
\caption{\textbf{Left}: Scatter plot of the distribution of geo-tagged Wikipedia articles together with some images (\textbf{right}) matched to the articles shown as green dots on the left plot. The title of the Wikipedia articles are written under each image. Zooming-in is recommended for visualization.
}
\label{fig:scatter_plot_examples}
\end{figure*}

%\s{transition to satellite stuff here is not smooth}
%\s{the story in the next paragraph could be made more linear and simple}
Because satellite images are geolocated, i.e., they correspond to specific locations (and times), they can be paired with other geolocated datasets (e.g., OpenStreetMap~\cite{kaiser2017learning}), exploiting spatial co-occurrence statistics 
%over space (and time) 
as a source of supervision~\cite{sheehan2018learning,sheehan2019predicting}.
%One recent study \cite{kaiser2017learning} utilizes OpenStreetMap building footprints to pre-train a convolutional segmentation network. They focused on four cities with highly accurate OSM footprints. After pre-training, they fine-tune the network on a dataset with perfect ground truth from the fifth city. Their experiments indicate that pre-training using OSM maps on satellite images is more beneficial than pre-training on Pascal-VOC07 dataset with perfect ground truth. 
%Following recent studies on %pre-training on satellite images, 
Following this strategy, we construct a novel multi-modal dataset by pairing  geo-referenced Wikipedia articles with their corresponding satellite images. By treating an article as an information-rich 
%\neal{hyphenate everywhere} 
label, 
we obtain highly detailed physical and qualitative context for each image. 
%to learn how to interpret the images in a CNN.  
For example, the first sentence of the John. F. Kennedy International Airport article 
%\footnote{\url{https://en.wikipedia.org/wiki/John_F._Kennedy_International_Airport}}
contains excerpts such as ``\textit{JFK is the primary international airport serving New York City}''. Wikipedia articles additionally contain demographic, environmental, and social information in structured form.
%which can be useful for other computer vision tasks. 
%highlighting the amount of information contained in only one sentence. 
%By utilizing this rich textual information in the CNN training pipeline, we remove the need to task human annotators to label 888,696 images, which is costly and time-consuming considering the difficulty of interpreting aerial images. 
To the best of our knowledge, this is the first time that Wikipedia has been used in conjunction with satellite images, and with 888,696 article-image entries, our approach yields the \emph{largest satellite image dataset} to date. 

%Wikipedia articles additionally contain demographic, environmental, and social information in structured form which can be useful for other computer vision tasks. Another exciting dimension that can be utilized is the accessibility of Wikipedia articles over time, as using alterations in article content and distribution over time for a region could be helpful in detecting changes and predicting economic or population growth, etc. We believe that the scale, coverage, and richness of this novel combination of crowdsourced annotations and satellite images will enable new advances in computer vision and a wide range of new applications. %\evan{Should we cite our KDD poverty paper here as proof?}
%\burak{We do not have it yet, and we should not cite our arxiv paper either. We can cite it later if the KDD paper get accepted.}
% To the best of our knowledge, this is the first time that Wikipedia has been used in conjunction with satellite images, and with 888,696 article-image entries, our approach yields the \emph{largest satellite image dataset} to date. By treating an article as an information rich label, we obtain highly detailed physical and qualitative context to learn how to interpret the images in a deep network.
%Additionally, demographic, environmental, and social information is often readily available in structured form in many Wikipedia articles. 

In this paper, we demonstrate the effectiveness of pairing Wikipedia articles to satellite images for pre-training CNNs for satellite image recognition.
%
%\s{explain explicitly what we do: extract simple tags, train a network to predict tags from images, transfer the weights, demonstrate they improve accuracy. for simplicity, we focus on africa. can use some of the text below}
We propose two pre-training methods to learn deep representations. First, similar to \cite{mahajan2018exploring},
we weakly label satellite images with curated summarization tags extracted from the article via an automated process. We then train a deep convolutional network to predict these weak labels directly from the images, learning useful representations in the process.
%Using coupled tags and images we pre-train a deep convolutional network. 
In the second approach, we propose a novel joint architecture where we first obtain a textual embedding of each article using document summarization techniques from NLP \cite{le2014distributed}
%\s{cite doc2vec} 
and then train a deep convolutional network to produce an embedding for each image that is ``similar'' to the textual one. The first approach is a crude way of getting a single weak label for each article whereas the second learns representations without weak labels. The pre-trained networks are then evaluated on a downstream hand-labeled dataset, as in \cite{jean2018tile2vec}, where we obtain $4.5\%$ higher accuracy compared to networks pre-trained on ImageNet, the standard approach for computer vision tasks.
%on satellite images.
% We believe this novel combination of visual and textual information will enable new applications for research in the social sciences, economics, sustainability, etc., via machine learning, computer vision, and natural language processing. In particular, it will complement existing data sources from surveys and open data portals such as Open Street Maps (OSM), which typically lack global coverage and provide more coarse information. % and lack temporal structure. 
%\neal{Does Wikipedia have temporal structure? This wasn't clear from the intro.}

\section{Pairing Rich Crowdsourced Annotations from Wikipedia to Satellite Images}
\label{sect:dataset}
%\subsection{Motivation}
%\s{start explaining the idea.}
%\s{wikipedia is a large scale crowdsourced bla bla with xxx articles in xxx languages....}
%\s{articles contain text, tables, etc}
%\s{out of x articles, y are geolocated, meaning we are given lat long coordinates}
%\s{give an example (perhaps using figure 2), but maybe too small to see the lat long field}

% \begin{figure}[h]
% \centering
% \includegraphics[width=0.50\textwidth]{./figures/article_example.png}\\
% \caption{An example Wikipedia article, with its coordinates boxed in red on the right-hand side. Observe the physical details about the bridge included in the article as well. Figure \ref{fig:labeling_examples} contains this article's paired ``Bridge'' image.}
% \label{fig:article_example}
% \end{figure}

Wikipedia is a large-scale, crowdsourced database spanning 302 languages
%\footnote{https://meta.wikimedia.org/wiki/List\_of\_Wikipedias} 
with over 47 million articles~\cite{wiki:gen}.
%for numerous topics, entities, figures, events, etc
Of these 47 million articles, about 11$\%$ are contained in the English version. 
%\footnote{https://en.wikipedia.org/wiki/Wikipedia:Size\_of\_Wikipedia}. 
%\neal{I think instead of footnotes we can probably just cite Wikipedia once.}
Out of these approximately 5 million
%\s{five?}
 articles, we found that roughly 1 million, or nearly 20$\%$, are geolocated, meaning there is a latitude and longitude $c_{i} = \{c_{i}^{lat}, c_{i}^{lon}\}$ associated with the article's text $y_{i}$. Our \emph{key idea is to use the article's coordinates to acquire a satellite image of its location from space} (see Fig. \ref{fig:scatter_plot_examples}).
%\s{numbers and percentages don't seem consistent}

%\s{i wonder if we should discuss text vs structured data (tables) in the articles}\burak{we can probably add a couple of sentences and explaining how it can be used to get climate, demography etc. information.}

%\s{having lat long allows us to acquire satellite image for each location.}
%\s{images are consistent: same size, viewpoint, etc.}
%\s{we need to say somewhere what kind of images we use, resolution, etc. either here or later}
%\evan{Burak does this in the "Images" section I believe.}
There is often a strong correlation between the article's text, $y_{i}$, and the visual content of the corresponding image, $x_{i}$.
Indeed, we can think of the article as an extremely detailed ``caption'' for the satellite image, providing an often comprehensive textual representation of the satellite image, or an \emph{information-rich label}. 
%\neal{Is ``data rich label'' an accepted term or are we just trying to use that throughout to refer to articles as labels? I feel like data rich is often used to refer to tasks with lots of labeled data --- something like ``information rich'' could be more appropriate?}
%\s{our key observation is that we can obtain sat. images for each location and that there is a relationship between the images and the article content}
%\s{then merge in some of the text below}
%\subsection{Parsing Wikipedia Articles}
%
%This creates the possibility for the utilization of extremely detailed, high-level, and descriptive image labels heretofore unavailable in the space. 
%
%Indeed, by pairing a Wikipedia article regarding some geolocated entity with a satellite image of its latitude and longitude, one can consider the article to be a detailed, often-comprehensive textual representation of the satellite image, or a data rich label. 
This label often contains structured data in the form of tables, called infoboxes, as well as raw text, allowing for the extraction of information about the physical state and features of the entity (e.g., elevation, age, climate, population).

\subsection{Acquiring Matching Satellite Imagery}
\label{sect:image_collection}
%\s{start by saying there are several ways to do the pairing, depending on the application. how large is the image, how many crops, center crops, etc. here we use the following process}
%\neal{Right now it feels like this subsection and the one in the previous section about satellite imagery are switched or at least need to be reorganized. In the previous section I think you can discuss different options for matching articles to satellite imagery and the tradeoffs involved. In this section, it seems like you want to concretely describe how you actually collected the image dataset that you use. Right now these are mixed together.}
%\s{agree with neal, that story/structure should flow reasonably well}
% Our novel multi-modal dataset relies on pairing 1.02 million articles to satellite images. 
%Previously, we discussed our pipeline to process Wikipedia articles. Now, we need to augment our dataset, WikiSatNet, with visual data. 
%\s{explain what you do. you pick an article, take the coordinates, acquire an image. there are multiple satellites, dates to consider, image sizes,. then explain the choices you did.}
%In the first step, we collected about 1 million geolocated articles globally from the latest Wikipedia dump release. In this section, we will detail how to pair an article, $x_{i}$, to a satellite imagery $y_{i}$. To build a pair, we first collect a satellite image centered at $c_{i}=\{c_{i}^{lat}, c_{i}^{lon}\}$, geolocation of article $y_{i}$. 
For a given article's coordinate $c_i$, there are many sensors that can provide imagery, with different tradeoffs in terms of spatial and temporal resolution, wavelengths, and costs. In this paper we acquire high resolution images from DigitalGlobe satellites. The images have a ground sampling distance (GSD) of $0.3$-$0.5$\textit{m}. These are among the highest resolution images available commercially, and 
%While high resolution DigitalGlobe images are not free, we prefer DigitalGlobe images since we can quantify our framework on 
were also used in 
the recently released 
%global, and large-scale 
functional map of the world (fMoW) dataset \cite{christie2018functional}.
%consisting of images from DigitalGlobe. 
Note that one could also use the same strategy to build a similar multi-modal dataset using lower-resolution (10 meter), publicly available Landsat and Sentinel-2 images. For a given coordinate $c_{i}$, there are usually multiple images available, captured at different times. We acquired the latest image available.
%as we consider the latest Wikipedia database dump from June 2018. 
%Thus, we can avoid potential conflict between the content of an article and its satellite imagery. This is particularly true for man-made objects, i.e, there can be a recently built airport that does not exist in images captured from earlier years. Next, we proceed with acquiring the images matching our strategy, however, 
%The fMow dataset can be interpreted as the ImageNet of satellite images. 
Another important design choice is the size of the acquired images. In this study, we use 1000$\times$1000 pixels images covering approximately an area of 900$m^{2}$. In aerial images, objects occupy drastically different numbers of pixels, as shown in Fig.~\ref{fig:scatter_plot_examples}. 
%For the airport case, larger image provides vital cues such as runway, plane, and terminal building \cite{christie2018functional} and 1000$\times$1000 pixels may not reveal enough information to interpret the object as airport as seen in Fig.~\ref{fig:scatter_plot_examples}, making the learning task harder. 
%On the other hand, for the stadium case, we can cover the object completely with a 1000$\times$1000 pixels image. An ideal approach would be to acquire larger image for airport and smaller for stadium articles, however, it is not practical to change the image size without human intervention. 
Based on preliminary manual examination, we found that 1000$\times$1000 pixels images can typically cover most of the relevant objects. Finally, we prioritized collecting RGB images and only acquired grayscale images if an RGB image was not available. We did not perform any filtering to remove cloudy images, as our goal is to learn robust representations on a noisy dataset.

Our resulting \emph{WikiSatNet} multi-modal dataset is a set of tuples $\mathcal{D} = \{(c_1, x_1, y_1), (c_2, x_2, y_2), \cdots, (c_N, x_N, y_N)\}$ where each tuple $(c_i, x_i, y_i)$ represents a location ($c_i$), corresponding DigitalGlobe image ($x_i$) and Wikipedia article text ($y_i$). \emph{WikiSatNet} contains
$N=888,696$ article-image pairs. To the best of our knowledge, this is \emph{the largest dataset to date consisting of satellite images} and about $2$ times larger than the recently released large scale fMoW dataset. 
%\s{this is x times larger than fmow}.
Note that our procedure is highly scalable and fully automated. It could be used to generate even larger datasets by considering other Wikipedia languages and other sensors in addition to DigitalGlobe.
%We call our dataset \emph{WikiSatNet} since it pairs satellite images and Wikipedia articles. 
In the next section, we propose two novel methods to pre-train a convolutional neural network (CNN) to extract information about images $x_{i}$ using information from  $y_{i}$.

\section{Learning Visual Representations using Wikipedia Textual Information}
Exemplifying the diverse application possibilities highlighted in the previous sections, we construct a general Wikipedia article-satellite image framework for pre-training CNNs. 
%using a weak supervision \cite{ratner2017snorkel} and unsupervised learning method \cite{jean2018tile2vec}. 
We then explore whether we can learn to interpret satellite images using knowledge extracted from Wikipedia articles via two approaches: weakly-supervised~\cite{ratner2017snorkel} labelling and a novel textual embedding method that attempts to match textual and visual embeddings.

\subsection{Weakly Supervised Learning}
%\s{this section is not particularly deep. try to cut down to about one third or half the current length.  it seems to me that if we said "we used regular expression matching to come up with labels and removed some" the reader wouldn't lose much..add some more detail about the most interesting/novel bits, and comment out the rest. we can add a sentence saying that scripts used will be released for reproducibility.}

%\burak{This section will be re-written by Evan}
We first propose learning visual features 
%(with a convolutional neural network) 
using a data-programming pipeline~\cite{ratner2017snorkel} to label our dataset. We begin by extracting a weak label $\hat{w}(y_i)$ for each article $y_i$ in our dataset. In our context, a weak label is a noisy, machine-generated classification of an article from a set of pre-defined labels.
%which consists of $888,696$ geolocated Wikipedia articles whose locations we possess a satellite image of. 
%we first need to engineer a standardized, interpretable labeling system.
Because of space constraints, we only provide a high-level description of the approach, and will add more details by purchasing extra pages in the final version.
As a first step, we manually compile a list of 97
%\s{how many?}
potential categories that an article could fall under (e.g., $city$, $lake$, $event$, etc.)
%\s{(e.g., xxx and yyy)} 
and use regular expressions to search for the terms throughout specific areas of the article's text where article meta-data is contained. We then rank the categories which are matched to the article in a manually-constructed hierarchical fashion from specific to general (e.g., $building \longrightarrow town \longrightarrow county$, etc.) and choose the one which comes first to label the article. 
%\s{the hierarchy is manually constructed and WHAT IS THE PRINCIPLE here? from more specific to more general?}
Because many of these category labels are very detailed, we then merge certain similar categories together to create more general labels. We also discard articles that are assigned labels which cannot be determined from a  satellite image 
%\s{rephrase} 
(e.g., $person$, $event$, etc.). Weak labels represented by less than 100 samples are also removed, reducing the final set of labels to 55. 
%We will release the scripts that perform these operations upon acceptance. 

%\s{if allowed, add an appendix, move the previous detailed text there, and say in the main text that details will be provided in appendix}
%\burak{I checked with the editor. It is not allowed.}
\begin{figure}[!t]
\centering
\hspace{-0.8cm} \begin{tabular}{@{}c@{}}North Queensland \\ Cowboys \end{tabular} \hspace{0.8cm} \begin{tabular}{@{}c@{}}Highland \\ Aviation \end{tabular} \hspace{1.2cm} Iserbrook
\includegraphics[width=0.47\textwidth]{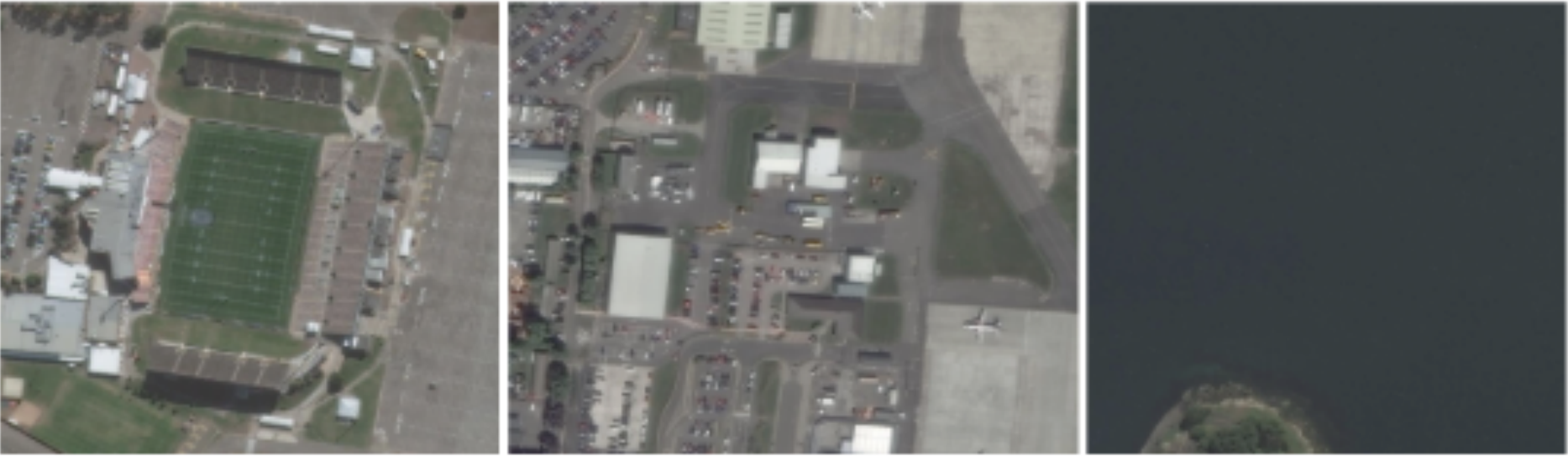}\\
\vspace{-0.15cm}
\hspace{0.3cm} (event) \hspace{1.5cm} (school) \hspace{1.5cm} (incident) 
\vspace{-0.20cm}
\caption{Some of the extracted weak labels representing \emph{flipped} label noise. Corresponding Wikipedia article titles are written above the images. Though the words \emph{stadium}, \emph{airport}, and \emph{water} are mentioned 19, 6, and 23 times in the articles, our weak label extraction pipeline generates wrong labels. Using image to text matching helps alleviate this flipped label noise.
%, we obtain vector representations of the articles.
%summarize articles, highlighting frequently mentioned words. 
%We then use the image to text matching to learn article summaries rather than weak labels to alleviate flipped label noise.
}
\label{fig:flipped_label_noise}
\end{figure}

\begin{figure}[!t]
\centering
\includegraphics[width=0.47\textwidth]{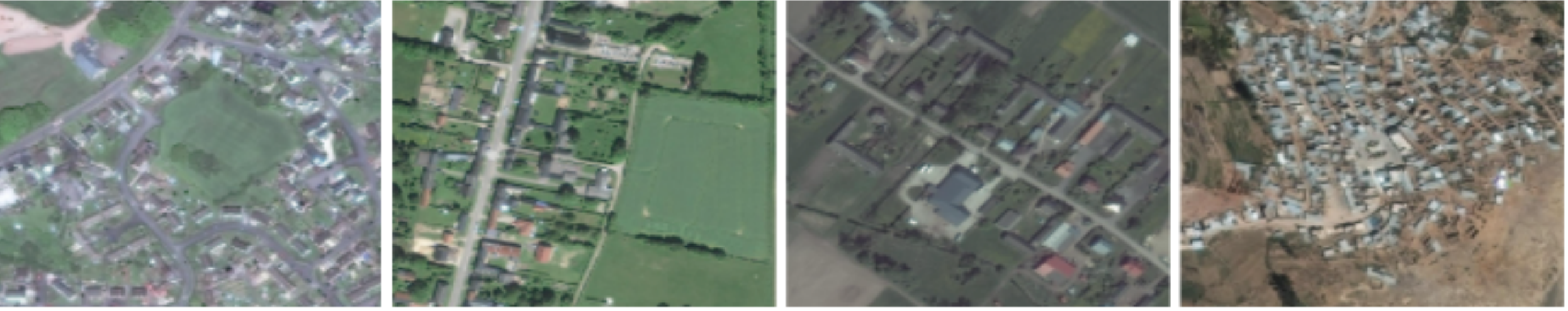}\\
\vspace{-0.15cm}
\hspace{0.2cm} (city) \hspace{1cm} (town) \hspace{1cm} (town) \hspace{1cm} (county) 
\vspace{-0.20cm}
\caption{Visually similar examples where the extracted weak labels cause \emph{adversarial} label noise. Here the CNN is penalized for errors even when the predicted label is visually similar to assigned weak label. In contrast, our document summarization model projects the embeddings of the articles of these images to a similar space to avoid penalizing the CNN when predicting a similar label.}
\label{fig:adversarial_label_noise}
\end{figure}

\begin{figure*}[!h]
\centering
\includegraphics[width=0.98\textwidth]{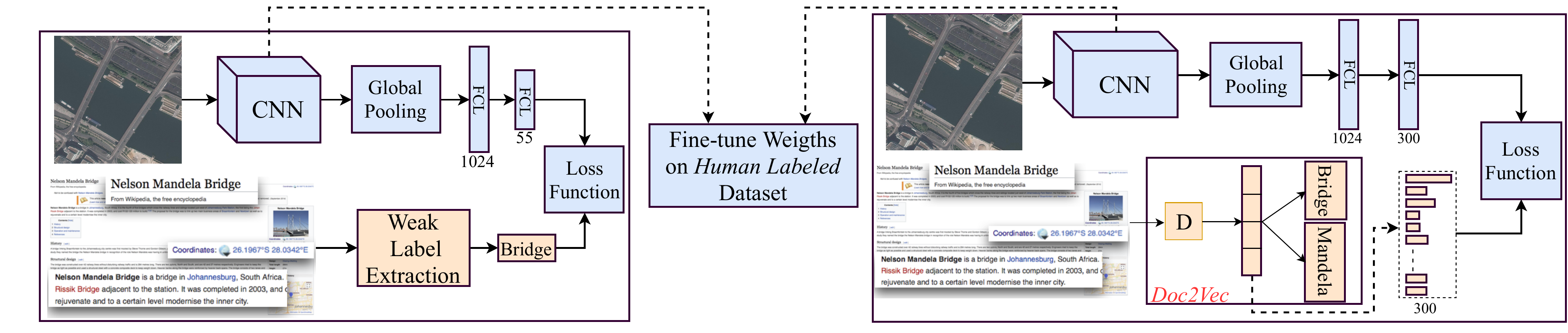}\\
\vspace{-0.15cm}
(a) \hspace{10cm} (b)
\vspace{-0.2cm}
\caption{The workflow of the proposed weakly supervised learning method (a): (1) Extract labels from articles using our labeling pipeline. (2) Match articles with images of their coordinates. (3) Pre-train on a large-scale dataset using 55 weak labels. (4) Transfer learned weights to a down-stream task. In (b) we show the workflow of the image to text matching learning. Our method enforces the CNN to learn features similar to raw textual features learned by \textit{Doc2Vec}.}
\label{fig:combined_workflow}
\end{figure*}

Given the final set of weak labels and corresponding images, we train a classifier to predict $\hat{w}(y_i)$ from $x_{i}$. The classifier is composed of a convolutional neural network $f_{v}: \mathcal{X} \mapsto \mathbb{R}^{M}$ that embeds images into an $M$ dimensional feature space, followed by fully connected and softmax layers as shown in Fig.~\ref{fig:combined_workflow}a.
%\s{fully-connected and softmax layers}. \s{give brief details on the architecture}. \s{see figure..}  
In this study, we parameterize $f_{v}$ using the DenseNet121~\cite{huang2017densely} architecture which was previously shown to perform well across a range of tasks.  The classifier is trained using the cross entropy loss function. The features learned by the convolutional embedding $f_{v}$ on this large-scale pre-training task can then be transferred to downstream tasks, e.g., object detection or land cover classification.
% learn the function, $f_{v1}: x_{i} \mapsto \mathbb{R}^{M_{1}}$, as shown in Fig.~\ref{fig:combined_workflow}a as
% \begin{equation}
%     z_{t}^{i} = f_{v1}(x_{i})
% \end{equation}
% \begin{equation}
%     L(z_{t}^{i}, \hat{w_{i}}(y_{i})) = \hat{w_{i}}(y_{i})\log(p(z_{t}^{i}))
% \end{equation}
% where $L$ represents the cross entropy loss function and the weights for the function $f_{v1}$ are transferred to a down-stream task for fine-tuning with \emph{perfect ground truth}. The function $p$, on the other hand, denotes the classification and softmax layers.
%\evan{We should be more specific about $f_{v1}$ and $f_{v2}$ if we introduce them here.}

% \begin{figure*}[!h]
% \centering
% \includegraphics[width=0.85\textwidth]{./weak_labels_distribution.pdf}\\
% \caption{Visualization of cosine similarities learned by CNN. The cosine similarities between CNN embeddings and \textit{Doc2Vec} embedding are computed and overlaid on the images. The CNN learns to embed AT$\&$T stadium's image closer to the its corresponding article.}
% \label{fig:embedding_similarity_example}
% \end{figure*}

Extracting weak labels is a noisy process that leads to a significant number of \emph{flipped} labels as shown in Fig.~\ref{fig:flipped_label_noise}. 
%\neal{Why are they called ``flipped''? Also in the caption for Figure 2, perhaps make it more clear that the flipped labels are below the image. I didn't understand the caption when I looked at only the figures.} 
Additionally, the process leads to \emph{adversarial} label noise because of visually similar labels such as \emph{city, country, populated place, building, town} etc., as shown in Fig.~\ref{fig:adversarial_label_noise}. 
% \neal{Why is this kind of noise considered adversarial? Instead of merely harmful?} 
% \burak{It is called adversarial when using a noise distribution where labels are changed to other classes that are more likely to be confused with the true class. Cat-Dog - Adversarial, Cat-Car Flipped Label Noise. It is harmful because you penalize the model even if it chooses a visually similar class. That is why merging visually similar labels reduce adversarial noise, but then merging operation lead to class imbalance problem.}
One can apply a simple merging step to place such visually similar labels into a general category, e.g., \emph{populated place}. However, it leads to a class imbalance problem where almost $40\%$ of the dataset is dominated by populated places. Exploring the trade-off between adversarial label noise and class imbalance problems is a very time-consuming process due to the nature of working with a large-scale dataset. 
%\neal{What is the goal of this sentence?} 
%\burak{You can check my previous comment} 
For this reason, in the next section, we propose a novel, and practical method to learn deep representations using multi-modal data without manual pre-processing.

\subsection{Image to Text Matching Learning}
%\evan{Needs a new title. I think Matching and Learning don't sound good next to each other. Then it needs a new name in the literature XD}
%\burak{Why? That is how it is called in the literature.}
In this section, we propose a novel method to learn deep convolutional features without using hand-crafted labeling functions. This not only substantially reduces human effort, but also tackles the adversarial label noise by softening the loss function for the images that can fall into multiple visually similar categories. Our method relies on the idea of image to text matching~\cite{lei2015predicting,wang2018learning}. In this direction, we propose a novel network shown in Fig.~\ref{fig:combined_workflow}b with two branches: a \emph{visual} and a \emph{textual} one. 
We design a loss function that encourages the CNN (\emph{visual} branch) to produce image embeddings that are close to a suitable vector representation of the corresponding article's text (\emph{textual} branch). 

The proposed architecture uses satellite images, $\mathcal{X}$, and Wikipedia articles, $\mathcal{Y}$, as input. In the \emph{textual} branch, we learn a function $f_{t}: \mathcal{Y} \mapsto \mathbb{R}^{K}$, to project an article, $y_{i}$, to a textual embedding space $z_{i}^{t} \in \mathbb{R}^{K}$ using a document summarization model from natural language processing (NLP):
%\neal{MLP?}\burak{It is NLP that produces a 1-D vector}
%\s{NLP network is not standard terminology. document summarization maybe?}
%\s{these formulas should have an $i$ subscript on the LHS. consider moving $t$ and $v$ as superscripts} \burak{Should be good now}
\begin{equation}
  z_{i}^{t} = f_t(y_{i}).
\end{equation}
%where $i$ represents the index of the Wikipedia article in $Y=\{y_{0}, y_{1}, ..., y_{N}\}$ and $N$ is the number of geo-tagged articles. 
In the \emph{visual} branch, we use a function $f_{v}: \mathcal{X} \mapsto \mathbb{R}^{M}$ parameterized using a convolutional neural network to extract features from an image as
%\s{for symmetry, give more details on the architecture of $f_{v}$, is it the same as in the previous section?} 
%\burak{Yes, it is the same. We mention it at the end of this section}.
%\s{notation is bad, get red of unnecessary subscripts}
\begin{equation}
  z_{i}^{v} = f_{v}(x_{i})
\end{equation}
where $i$ represents the index of the image paired to article $y_{i}$. We parameterize $f_{v}$ using the DenseNet121 architecture \cite{huang2017densely} as in the weak supervision method. Next, we use a function $f_{m}: \mathcal{Z}^{v} \mapsto \mathbb{R}^{K}$ to map $z_{i}^{v}$ to the same dimension as the textual feature vector $z_{i}^{t}$. The function $f_{m}$ is parameterized using a fully connected layer with ReLU activations.
%\s{you haven't defined/mentioned it yet..}.
%\burak{I swapped the order. Makes more sense to me now.}
% \begin{equation}
%   z_{v} = f_{m}(z_{v}).
% \end{equation}
%\neal{I don't think this notation makes sense --- if you take the L2-norm of a vector then you get a scalar right? Double check this?}
%The final feature vector, $z_{v2}^{i} \in \mathbb{R}^{M_{2}}$, matches the dimension of the textual feature vector. 
% On the other branch, we learn the function, $f_{t}: y_{i} \mapsto \mathbb{R}^{K}$, to project the article, $y_{i}$, to a textual embedding space $z_{t} \in \mathbb{R}^{K}$ through a document summarization model as 
% %\neal{MLP?}\burak{It is NLP that produces a 1-D vector}
% %\s{NLP network is not standard terminology. document summarization maybe?}
% \begin{equation}
%   z_{t} = f_t(y_{i}).  
% \end{equation}
The final feature vectors, $z_{i}^{v}$ and $z_{i}^{t} \in \mathbb{R}^{K}$, are then compared with a loss function that enforces similarity.

\subsubsection{Pre-training the \textit{Doc2Vec} Model}
\begin{figure}[!h]
\centering
\subfloat{\includegraphics[width=0.25\textwidth]{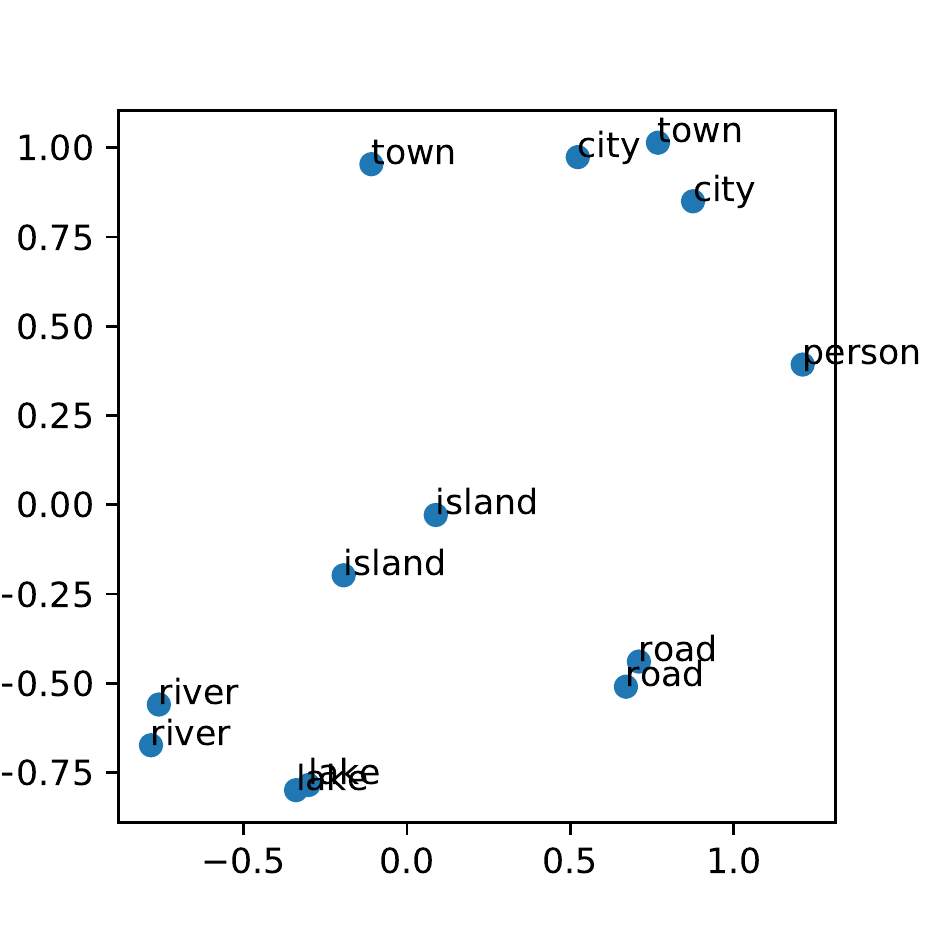}}
\subfloat{\includegraphics[width=0.25\textwidth]{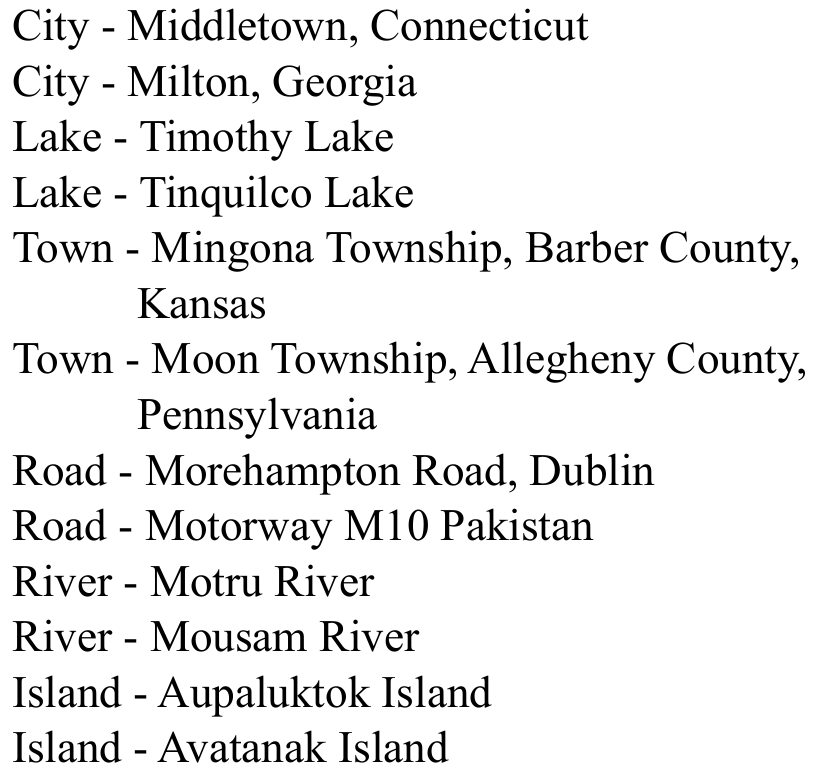}}
\caption{Visualization of PCA components of the randomly chosen articles learned by \textit{Doc2Vec}. Notice that visually similar objects such as \emph{city}, \emph{town} are closely mapped while different objects are projected far away. Corresponding Wikipedia article titles are shown on the right. 
%\evan{Person article title?}
}
\label{fig:textual_embeddings}
\end{figure}

Our image to text matching method uses textual descriptors $\mathcal{Z}^{t}$ to learn deep visual representations. 
%Recently, numerous approaches, including Bag of Words and TF-IDF, have proposed learning textual descriptors from a document \cite{klein2014fisher}. 
In our study, we use the \textit{Doc2Vec} network \cite{le2014distributed} which can summarize variable length articles in a unified framework. \textit{Doc2Vec} is a document summarization method that can take a variable length piece of text, $y_i$, and map $y_i \in \mathcal{Y}$ to a paragraph vector $z_{i}^{t}=f_t(y_i) \in \mathbb{R}^K$ in a fixed-length vector space, where $K$ is specified by the user. Documents that possess similar meanings are mapped to nearby points in the embedding space, allowing a comparison between any two documents. In contrast to fixed length vector representations using Bag-of-Words, \textit{Doc2Vec} can capture the orderings and semantics of the words, which is highly beneficial for our unsupervised learning task. For example, learning a textual embedding space where we can closely map article categories such as \emph{country}, \emph{city}, \emph{town} etc. is desired considering that their corresponding visual data contain similar structures (see Fig.~\ref{fig:textual_embeddings}).
%\evan{Should this say map articles about \emph{countries}, \emph{cities}, \emph{towns}, etc. to a similar space instead of mapping the words themselves?}
Another advantage of the \textit{Doc2Vec} model is that it is an unsupervised learning model. This allows us to learn Wikipedia-specific descriptors by training it on the full geolocated Wikipedia article corpus. 
% The original paper proposes two training frameworks;(1) learning using vector representation of the words, and (2) learning using distributed bag of the words (DBOW). In this study, we learned $f_{t}$ using the DBOW approach for being a more efficient method as the first one needs to learn and store the word vectors. After training \textit{Doc2Vec} on Wikipedia articles, we freeze its weights on the image to text matching learning task.

\subsubsection{Cosine Similarity Loss Function}
After learning feature vectors, $z_{i}^{v}$ and $z_{i}^{t} \in \mathbb{R}^{K}$, from the two branch network, we apply a loss function to measure the similarity of the two vectors. We propose using the cosine similarity metric, which measures the angle, $\theta_{i}$, between two vectors as
\begin{equation}
    D(x_{i}, y_{i}) = cos(\theta_{i}) 
     = \frac{f_v(x_i)^T f_t(y_{i})}{\left\Vert f_v(x_i)\right\Vert_2 \left\Vert f_t(y_{i})\right\Vert_2}.
    %= \dfrac{\sum_{k=1}^{K} z_{t}^{k} z_{v}^{k}}{\sqrt{\sum_{k=1}^{K} (z_{t}^{k})^{2}} \sqrt{\sum_{k=1}^{K} (z_{v}^{k})^{2}}}
    % \dfrac{z_{t}^{i}.z_{v2}^{i}}{||z_{t}^{i}||\:||z_{v2}^{i}||}
\end{equation}
%\s{formula seems wrong}
%where $K$=$300$ represents the dimension of the embedding space from text and image branch.

Wikipedia has varying lengths of articles, which makes the cosine similarity function ideal since it measures the similarity between the direction rather than the magnitude of two vectors.
% \evan{not sure article length corresponds to magnitude.}
% \burak{Yes, it does. Having one word and five words of AI in a short and long article should be treated same. In other words, the first one is not less related to machine learning than the second one.}\evan{yes i agree. what i mean is that i don't think doc2vec inherently gives different magnitudes to different length articles.}
%\evan{Don't the L2 layers set the magnitude of both vectors to 1 already though?}
%\burak{That is a good point. I kept it to have flexibility to use other loss functions, but it can be removed.}

%The proposed image to text matching method as shown in Fig.~\ref{fig:combined_workflow}b can be directly used on the 888,696 images and articles. On the other hand, our weak supervision method as shown in Fig.~\ref{fig:combined_workflow}a relies on a number of post-processing steps to determine a final set of weak labels, reducing the number of articles to 725,559.
One can apply some other loss functions for our pre-training task. For example,  \cite{wang2018learning} proposed triplet loss function where the anchor is a phrase paired to its corresponding positive visual data. The negative image is then sampled from the neighborhood of the positive sample. \cite{jean2018tile2vec} adapted triplet loss function for unsupervised learning on satellite images. They assign a positive image for each anchor image from its spatial neighborhood following the assumption that nearby images contain similar visual structures. The negative sample is then sampled from the areas outside the anchor's neighborhood circle. In our case, we lack explicit knowledge that can help us sample negative image given an article, $y_{i}$, as anchor and its corresponding image, $x_{i}$, as positive. In this direction, one can compute the similarity in the visual, $z_{v1}$, or textual, $z_{t}$, embedding space between a positive sample and other samples in a certain spatial neighborhood to get a negative sample.

Another interesting aspect of our architecture is the dimensionality of the textual embeddding space. We believe that 300 dimensional feature vector can capture all the fine-grained visual structures in an aerial image. However, during our experiments we observed that visually similar features can lead to more uniform textual descriptors slowing down the learning process. Thus, using a smaller dimensional embedding space can lead to more discriminative visual features that can potentially speed up the learning process. On the other hand, this can also prevent the CNN from learning fine-grained in- formation. We leave the task of exploring this trade-off as a future work of our study. Another future dimension of our image to text matching work is generating a Wikipedia article, $y_{i}$, using an NLP decoder $f^{'}_{t}$ given visual representations $x_{i}$ as 
\begin{equation}
    y_{i} = f^{'}_{t}(f_{v}(x_{i})).
\end{equation}

\subsubsection{Training on \textit{WikiSatNet}}
%\s{we train the architecture on our dataset xx using the optimizer yy and initialization xxx}
%\s{i feel this subsection could be expanded a bit, it's one of the main contributions of the paper}
%\burak{I think this paragraph should go into the pre-training section in 4.1}
%\s{no this should go in the learning representations section}
%After proposing two pre-training methods, we now detail our experiments using \textit{WikiSatNet}.
In our pre-training experiments, we use similar hyper-parameters in both weak supervision and image to text matching to train the DenseNet121 for optimizing the weights for $f_{v}$. We initialize weights randomly, however, we observed faster convergence when initializing with pre-trained weights.
%Starting with a model pre-trained on ImageNet is helpful in reducing the optimization time for both pre-training methods. \s{this feels like cheating. if we use imagenet as pre-training, then they might wonder if this is the reason the method works at all}
After experimentation, we set the learning rate and batch size to 0.0001 and 64, respectively, and the Adam optimizer is used to train the model \cite{kingma2014adam}. Finally, we resize the 1000$\times$1000 pixels images to 224$\times$224 pixels images to compare with publicly available datasets. 
%\neal{A bit confused here, doesn't resizing the images change the resolution?}
%\burak{Sure it does. How are you going to fine-tune the model on 224x224 images if you pre-train them on 1000x1000 images? Besides, you will not be able to compare to ImageNet.}

\begin{figure}[!h]
\centering
\includegraphics[width=0.47\textwidth]{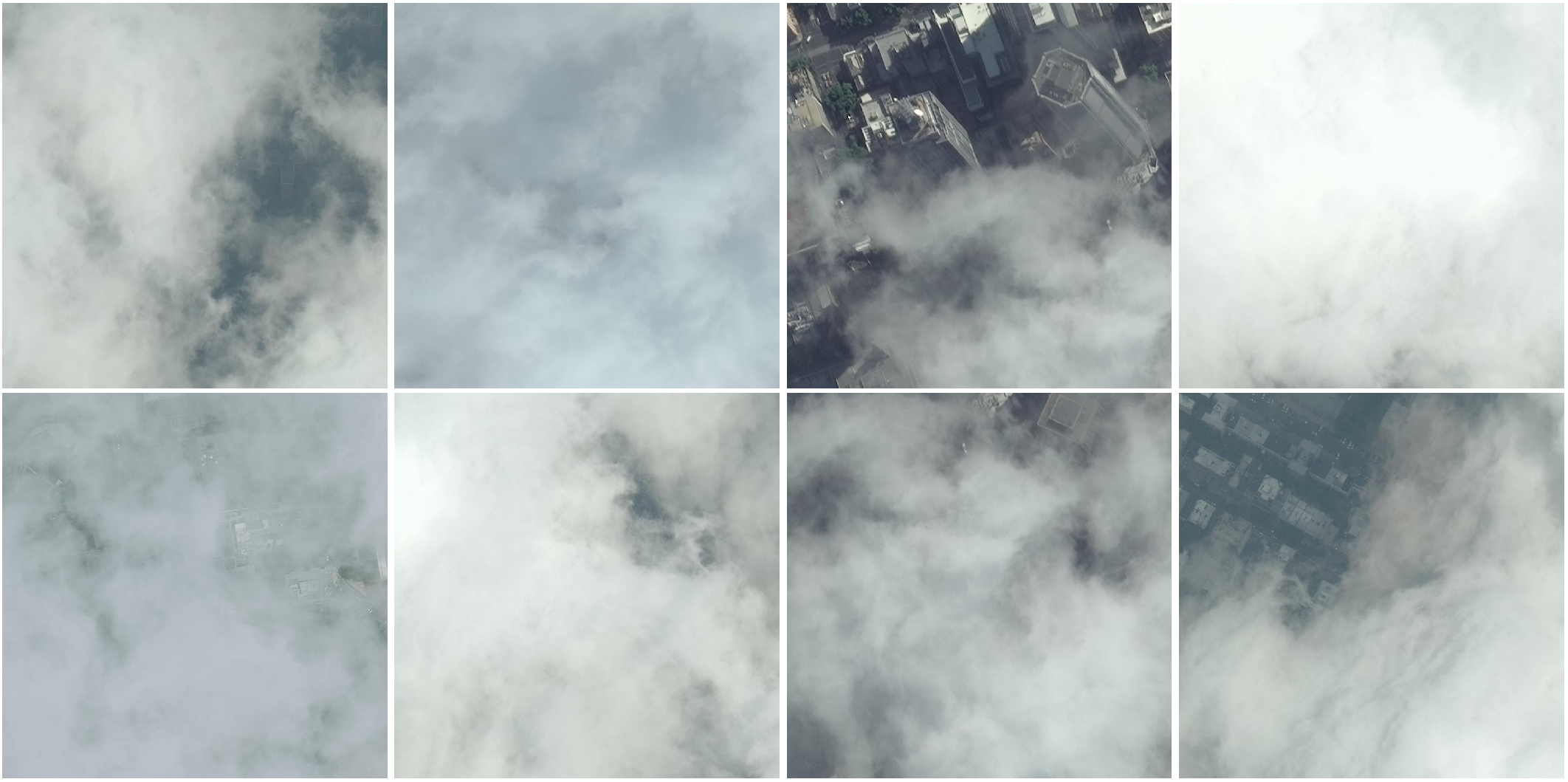}
\caption{Some of the cloudy images in WikiSatNet. The cloudy images amount to roughly $5$-$7\%$ of the dataset.}
\label{fig:cloudy_images}
\end{figure}

\begin{figure}[!h]
\centering
\includegraphics[width=0.45\textwidth]{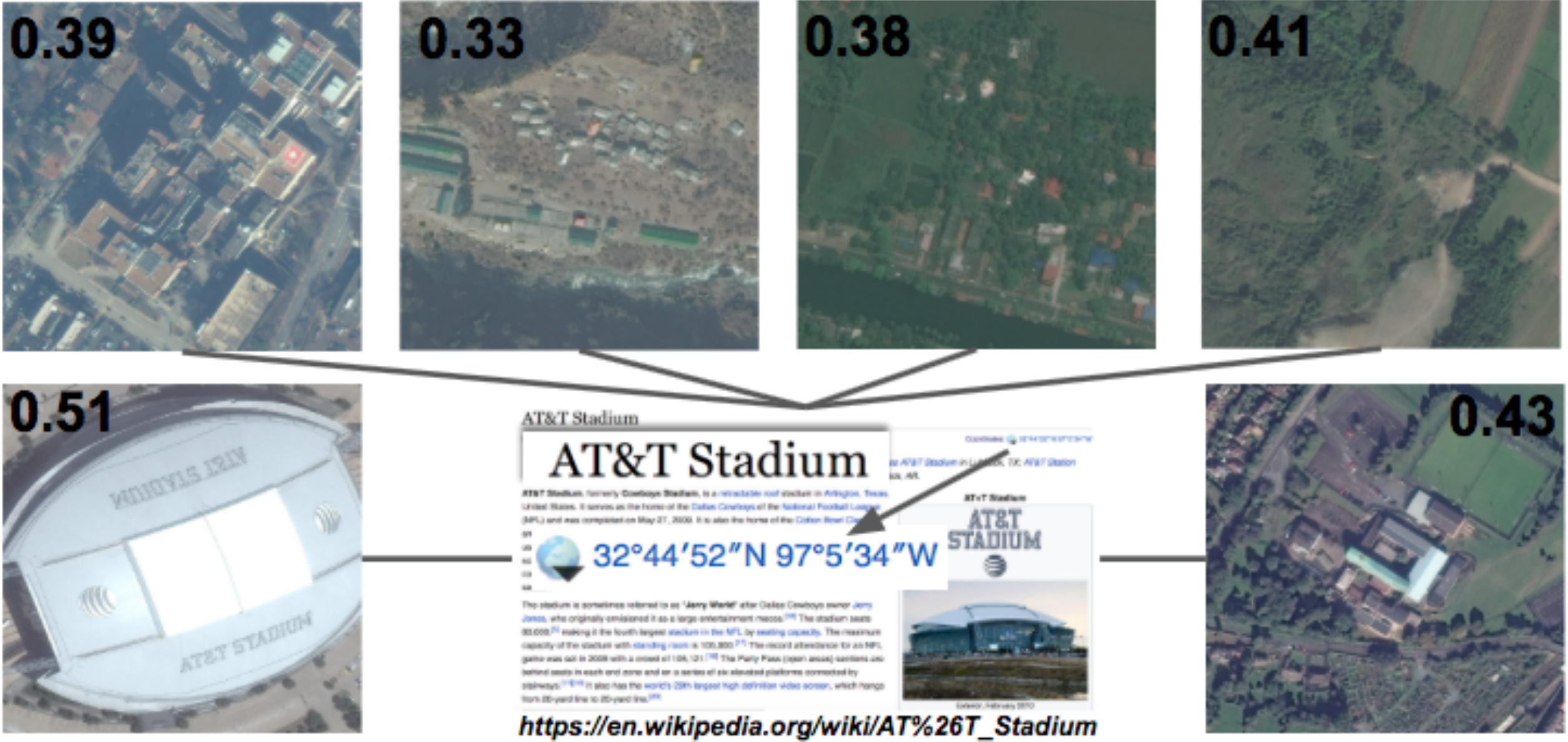}\\
\caption{Visualization of cosine similarities learned by the CNN. The cosine similarities between the CNN embeddings and the \textit{Doc2Vec} embedding are computed and overlaid on the images. The CNN learns to embed AT$\&$T Stadium's image closer to the its corresponding article.}
\label{fig:embedding_similarity_example}
\end{figure}

In the initial steps of image to text training, we observe an angle of approximately $90^{o}$ ($D(x_{i},y_{i})\approx0$) between $z^{t}_i$ and $z^{v}_i$. This is consistent with the fact that random vectors in high dimensional spaces are likely to be orthogonal to each other. %\neal{On average for each minibatch? Would be kind of interesting to see how this distribution changes as you train.}
%\burak{Yes, on average for each minibatch.}
After several epochs, the angle decreases to about $45^{o}$ ($D(x_{i},y_{i})\approx0.5$) and stops decreasing further. 
%We observed the same trend with the weak supervision experiments. 
We believe that this is partially due to articles that do not contain any visual cue, e.g \emph{culture} and \emph{person}, and also cloudy images (see Fig.~\ref{fig:cloudy_images}), which amount to roughly $5\%$ of the dataset. 
%Meanwhile, for the weak supervision, it might be due to \emph{flipped} and \emph{adversarial} label noise and \emph{cloudy} images.
We did not observe over-fitting in our experiments.
While we are not able to achieve zero loss, we qualitatively find that our approaches learn meaningful representations. 
To verify this, after pre-training the CNN on \textit{WikiSatNet} using the image to text matching, we visualize the cosine similarities between $z_{i}^{t}$ and $z_{i}^{v}$ as shown in Fig.~\ref{fig:embedding_similarity_example}. In the same figure, we keep $z_{t}$ fixed and use embeddings from images at different locations. The CNN learns to project embedding $z_{i}^{v}$ closer to its corresponding article embedding $z_{i}^{t}$. We will publicly release the code for our image to text matching and weak supervision methods upon publication. Additionally, we expect to release a substantial fraction of the high resolution images in \textit{WikiSatNet} (negotiations on the license with the image provider are ongoing).
%that ensures performance boost on the image recognition task on the fMoW. 
This will encourage further research into jointly utilizing Wikipedia and satellite images.
%\neal{Consider adding some paragraph breaks in this section.}

\section{Transfer Learning Experiments}

After pre-training CNNs on \textit{WikiSatNet} using the proposed methods, we test them on three target tasks: (1) single image classification on the fMoW dataset, (2) temporal view classification using multiple images over an area on the fMoW dataset, and (3) land cover classification. In these tasks, we compare our pre-training strategies to the following baselines: (1) pre-training on ImageNet \cite{russakovsky2015imagenet}, (2) pre-training on CIFAR10, and (3) training from scratch.
%on fMoW. 
Our goal is to evaluate whether we learn 
satellite-specific representations that outperform the ones obtained using 
%publicly available, 
out-of-domain benchmarks with human labels. 
%\s{not sure what the previous sentence means}
% The ImageNet dataset comes from the Imagenet Large Scale Visual Recognition Challenge that contains about 1.2 million training images from 1000 word synsets. The CIFAR10 dataset, on the other hand, contains 50000 training images across 10 categories. By using two image recognition datasets with different scales, we can quantify the effect of the size of out-of-domain pre-training dataset on the target task consisting of aerial images.

\subsubsection{Fine-tuning}
%We use the DenseNet121 model pre-trained on the WikiSatNet and transfer the weights for classification on the fMoW dataset. 
There are two classical approaches in fine-tuning a deep network on the target task: (1) training all layers, and (2) freezing all the layers other than the final classification layer. In our experiments, we present results from both strategies.  %\neal{Capitalize table and figure refs.} 
%The classification layer of the model, on the other hand, is changed to perform 62-way classification to match the number of labels in fMoW. 
The learning rates for the weakly supervised and image to text matching model are set to 1e-4 and 1e-5 after experimentation. On the other hand, the learning rate for the ImageNet model is set to 1e-4, while it is set to 1e-3 for both the CIFAR10 pre-trained and trained from scratch models. These were the best performing hyper-parameters in our experiments. 
%\s{how were these learning rates chosen?} \burak{By experimentation.}
%\s{can we say these were the best performing ones?}
%\burak{Yes, we can say that}
Finally, resized 224$\times$224 pixel RGB images are used as input to the model as in the pre-training task. 
%\s{i thought pre-training was 1000x1000?. again, make that clear in the pre-training part. do not mix pre-training and fine-tuning}
%\burak{I tried to make it more clear. We use 224x224 images as Input to the model in all the stages.}
We follow the same approach for the models pre-trained on CIFAR10 and ImageNet.

\subsection{Experimenting on the fMoW Dataset}

To quantify the quality of the  representations learned in the pre-training step, we first use a recently released large-scale satellite image recognition dataset named fMoW~\cite{christie2018functional}. The fMoW dataset consists of both multispectral and RGB images and contains 83,412 unique training bounding boxes from large satellite images representing 62 different objects. The validation and test sets contain 14,241 and 16,948 bounding boxes and are left unchanged in our experiments. It also comes with temporal views from the same scenes, making classification of some classes such as \textit{construction site} and \textit{flooded road} easier. \cite{christie2018functional} proposes a multi-modal architecture that uses a DenseNet161 pre-trained on ImageNet and an LSTM to learn from images and their corresponding meta-data. Their DenseNet161 model has a number of parameters similar to the DenseNet121 model we use in our experiments. Since our pre-training framework learns from visual data, it can be easily applied to any CNN model to boost performance as well as reduce the number of labeled samples needed for a target task. 
%\s{are we using their CNN or the densenet? this is still not clear}
%\burak{We are using DenseNet121, they are using DenseNet161. However, the number of parameters are similar. The performances should be similar.}

%subsection{Results} 

\subsubsection{Reasoning on Single Images}
\begin{figure}[!t]
\centering
\subfloat{\includegraphics[width=0.25\textwidth]{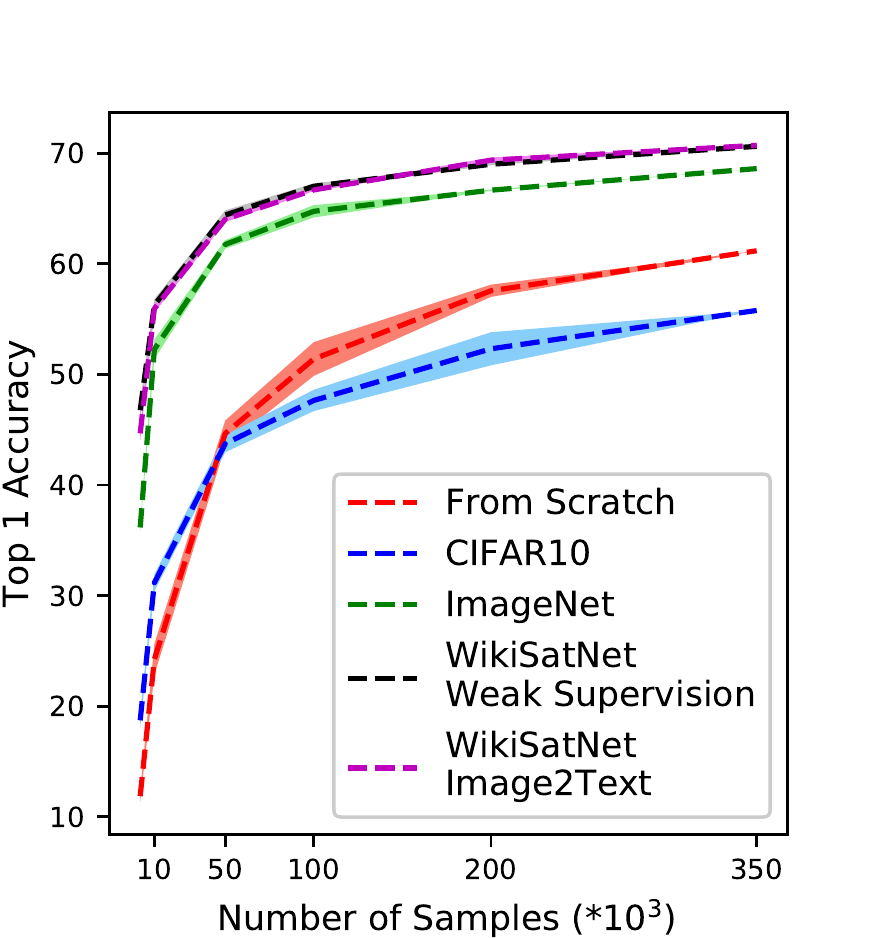}}
\subfloat{\includegraphics[width=0.25\textwidth]{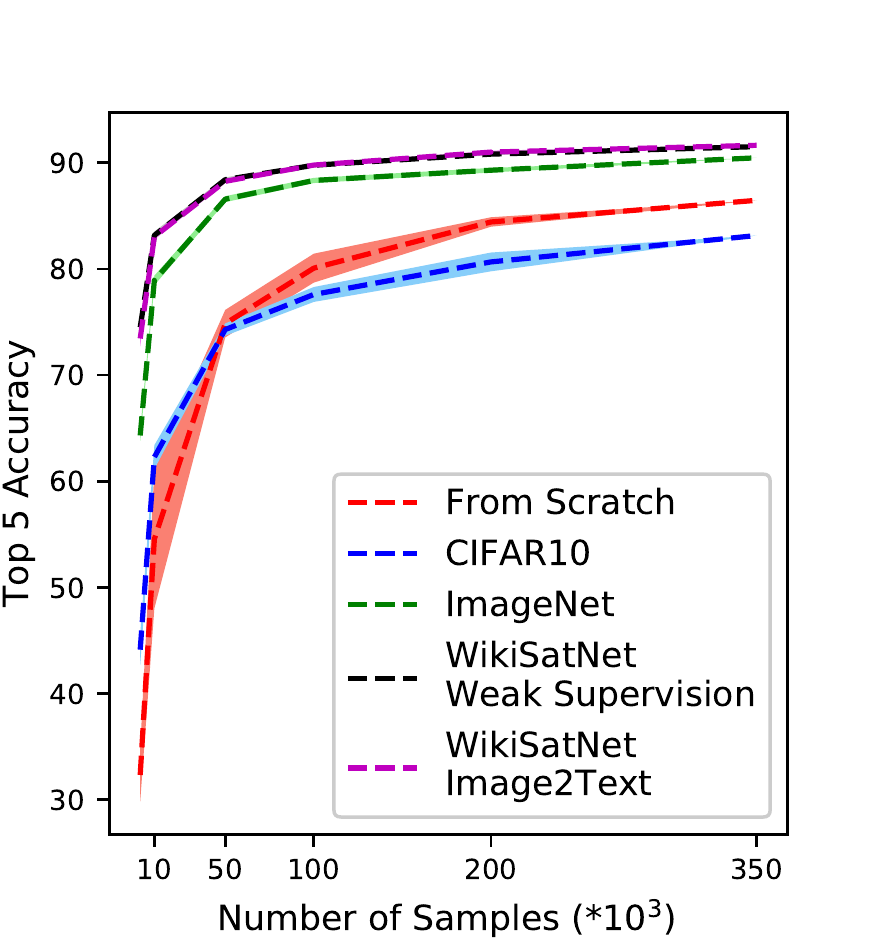}}
\caption{The top-1 and 5 classification accuracies of the proposed pre-training and baseline strategies on fMoW's test set when fine-tuning all layers on fMoW's training set. Monte-Carlo experiments were conducted when sampling a subset of the full training set. 
% Image to text matching method performs similarly to the handcrafted label extraction method whereas they both outperform other pre-trained strategies by large margin especially when sampling relatively small number of labeled samples. Even considering full training set (350000 samples), our pre-training strategies outpeform ImageNet pre-trained model. One can also observe that models pre-trained on WikiSatNet produces models with less uncertainty on the fMoW task.
}
\label{fig:single_image_reasoning}
\end{figure}

In the first task, we perform experiments on the fMoW dataset for the task of classifying individual images using features extracted by the \emph{visual} branch $f_{v}(\cdot)$ as 
\begin{equation}
    L(x_i) = \argmax_{j}(p(f_{v}(x_i)))
\end{equation}
%\s{i don't think this equation is necessary} \burak{We can remove it if we do not have any space.}
where $p$ represent the fully connected and softmax layers whereas $j$ denotes the index of the assigned label. We experiment with 2000, 10000, 50000, 100000, 200000, and 350000 training images. As shown in Fig.~\ref{fig:single_image_reasoning}, our pre-training strategies outperform the other pre-training strategies by large margins in top-1 and top-5 classification accuracy when using small amounts of labeled data. For example, when using 2000 labeled images, both our training strategies outperform ImageNet and CIFAR10 by $10\%$ and $30\%$, respectively. 
%\neal{This is for Top-1?} 
As expected, this number goes down to about $5\%$ and $20\%$ when increasing the number of labeled images to 50000. Interestingly, at this point, the model trained from scratch starts to outperform the model pre-trained on CIFAR10. When using the full training set, our proposed pre-training strategies outperform ImageNet by about $2\%$ and outperform the model trained from scratch by about $10\%$. These results demonstrate that our proposed approach produces 
%domain-specific %\s{what does domain specific mean here?} 
%\burak{specific to aerial images} 
features that are highly beneficial in down-stream tasks involving satellite images, even when large numbers of human labeled samples are available. 
%The performance boost, on the other hand, is larger when considering small amount of human labeled samples.  
When fine-tuning only the final layer, the proposed pre-training methods outperform ImageNet features by about $13\%$ on the test set as shown in Table~\ref{table:freeze_layers_results}.
%\s{use bold to highlight winning methods in tables}

\begin{table}[h]
\centering
\resizebox{0.49\textwidth}{!}{\begin{tabular}{@{}lcccc@{}}
Model & \begin{tabular}[c]{@{}c@{}} CIFAR10 \end{tabular} & \begin{tabular}[c]{@{}c@{}} ImageNet \end{tabular} & \begin{tabular}[c]{@{}c@{}} WikiSatNet \\ \textit{Weak Labels} \end{tabular} & \begin{tabular}[c]{@{}c@{}} WikiSatNet \\ \textit{Image2Text} \end{tabular} \\
\hline
\begin{tabular}[c]{@{}c@{}} Top-1 Acc. \\ (Fixed $f_{v}$) \end{tabular} & 13.98 (\%) & 37.73 (\%) &  50.73 (\%) & \textbf{51.02} (\%) \\
\begin{tabular}[c]{@{}c@{}} Top-1 Acc. \\ (Fine-tuned $f_{v}$) \end{tabular} & 55.79 (\%) & 68.61 (\%) &  70.62 (\%) & \textbf{70.72} (\%)
\end{tabular}
}
\caption{Top-1 accuracies on the fMoW test set for pre-trained models. All the models are fine-tuned on the full fMoW training set. Fixed $f_{v}$ represents the fine-tuning method where the pre-trained weights are fixed whereas the second method fine-tunes all the layers. 
%\neal{Consider using booktabs style tables}
}
\label{table:freeze_layers_results}
\end{table}

\subsubsection{Reasoning on Temporal Views}
% Previously, we explored the fMoW dataset for the task of single image classification to quantify domain-specific features learned through our pre-training methods. 
In this section, we evaluate our representations on the task of temporal view classification across 62 classes from the fMoW dataset. 
%\neal{This sentence is unclear} 
This way, we can understand if our pre-training methods also boost performance on tasks that use temporal data as input.
%\neal{This sounds a bit weird, maybe try: In this section, we evaluate our representations on the task of temporal image classification across 62 classes from the fMoW dataset. We want to see if our pre-training methods can also boost performance on tasks that use temporal data as input.}
\cite{christie2018functional} trains the network on single labeled images 
%with the labels representing the area of interest 
and at test time averages the softmax predictions of the network on different images from the same area to assign the label with the maximum average score. 
%\s{this previous sentence is unclear} 
%\burak{I hope it is more clear now}. 
We follow their training and test methods and at test time average predictions from $T$ images over the same area, again using features extracted from $f_v(\cdot)$ as input. This can be formulated as
\begin{equation}
    L(X) = \argmax_{j}(mean(\sum_{t=1}^{T}p(f_{v}(x_{t}))))
\end{equation}
where $j$ denotes the index of the assigned label and $f_{v}$ represents the pre-trained network fine-tuned on the fMoW. 
%\evan{we have already used $N$} 
Different from the previous section, we now report results in F1-scores to compare our models to the ones proposed by \cite{christie2018functional}.
\begin{table}[h]
\centering
\resizebox{0.48\textwidth}{!}{\begin{tabular}{@{}lcccc@{}}
Model & \begin{tabular}[c]{@{}c@{}} CIFAR10 \end{tabular} & \begin{tabular}[c]{@{}c@{}} ImageNet \end{tabular} & \begin{tabular}[c]{@{}c@{}} WikiSatNet \\ \textit{Weak Labels} \end{tabular} & \begin{tabular}[c]{@{}c@{}} WikiSatNet \\ \textit{Image2Text} \end{tabular} \\
\hline
\begin{tabular}[c]{@{}c@{}}F1 Score \\ (\textit{Single View}) \end{tabular} & 55.34  (\%)  & 64.71 (\%) &  66.17 (\%) & \textbf{67.12} (\%) \\
\begin{tabular}[c]{@{}c@{}}F1 Score \\ (\textit{Temporal Views}) \end{tabular} & 60.45  (\%)  & 68.73 (\%) & 71.31 (\%) & \textbf{73.02} (\%) 
\end{tabular}}
\caption{F1 scores of different pre-training methods on fMoW's test set when fine-tuning all the layers on fMoW's training set.}
\label{table:temporal_view_results}
\end{table}

\begin{figure*}[!h]
\centering
\includegraphics[width=0.95\textwidth]{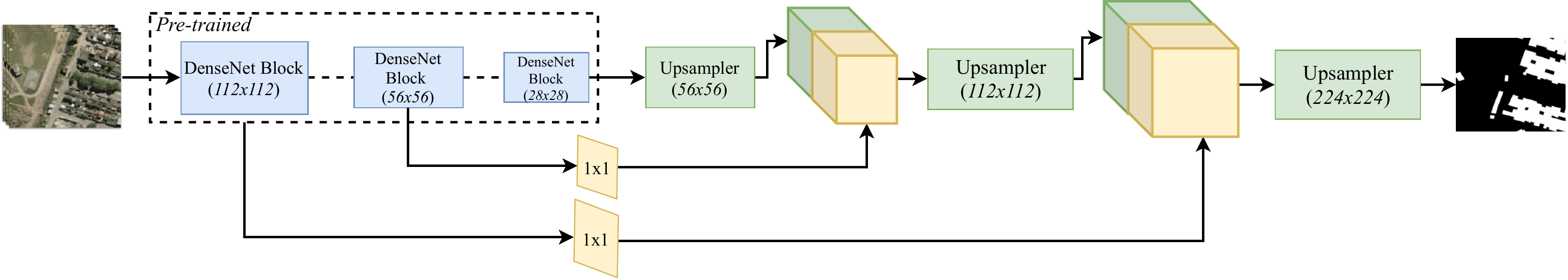}\\
\caption{The proposed segmentation architecture that uses the pre-trained weights in the encoder stage.}
\label{fig:segmentation_arch}
\end{figure*}
We first compare our pre-training methods to ImageNet and CIFAR10 pre-training in Table~\ref{table:temporal_view_results}. The proposed pre-training methods outperform the ImageNet pre-trained model by up to $4.5\%$ in F1 Score when performing reasoning on temporal views. Among the proposed methods, the image to text matching approach outperforms the weak supervision with handcrafted labels method by about $1.7\%$ in F1 Score. 
%\neal{Assuming that we're suddenly switching to F1 score so that we can compare directly to their results?} 
These results prove that the importance of pre-training does not diminish when switching from single to temporal views.
%, which provides more information to the decision process. 
On the other hand, \cite{christie2018functional} proposes five different models for the fMoW classification task. Three of them use meta-data and images jointly, whereas the remaining two only employ an ImageNet pre-trained DenseNet on images. Their visual data-only models are named \textit{CNN-I-1} and \textit{CNN-I}, where the former is a single view model and the latter performs temporal reasoning. We can improve these models with our pre-training strategy by about $4.5\%$ in F1 score while performing similarly to their top performing model, \textit{LSTM-IM}, which uses meta-data and visual data jointly to perform temporal reasoning. Although this is outside the scope of this paper, our pre-trained models can replace the DenseNet model, pre-trained on ImageNet, used in \textit{LSTM-IM} to improve its results as well.

Our experiments demonstrate that pre-training with weak or no supervision is very useful for the target task as reported by \cite{mahajan2018exploring} both in terms of (1) boosting accuracy and (2) reducing the required human labeled dataset size on the target task. Unlike the pre-training framework proposed by \cite{mahajan2018exploring}, we do not necessarily need to have billions of images to overcome noise to learn useful representations. 

\subsection{Experiments on Land Cover Classification}
Additionally, we perform classification across 66 land cover classes using remote sensing images with $0.6m$ GSD obtained by the USDA's National Agriculture Imagery Program (NAIP). We focus on the images from the California's Central Valley near the city of Fresno for the year 2016. The corresponding land cover map, named the Cropland Data Layer (CDL), is collected by the USDA for the continental United States \cite{NAIP2016}. The CDL is provided at $30m$ GSD, and we upsample them to match $0.6m$ GSD to use as ground truth. The final dataset consists of 100000 training and 50000 validation and test images. We only fine-tune the classification layer while keeping $f_{v}$ fixed.
\begin{table}[h]
\centering
\resizebox{0.48\textwidth}{!}{\begin{tabular}{@{}lcccc@{}}
Model & \begin{tabular}[c]{@{}c@{}} CIFAR10 \end{tabular} & \begin{tabular}[c]{@{}c@{}} ImageNet \end{tabular} & \begin{tabular}[c]{@{}c@{}} WikiSatNet \\ \textit{Weak Labels} \end{tabular} & \begin{tabular}[c]{@{}c@{}} WikiSatNet \\ \textit{Image2Text} \end{tabular} \\
\hline
\begin{tabular}[c]{@{}c@{}}Top 1 Acc. \end{tabular} & 42.01 (\%) & 40.11 (\%) & 46.16 (\%) & \textbf{47.65} (\%) \\
\begin{tabular}[c]{@{}c@{}}Top 5 Acc. \end{tabular} & 74.73 (\%) & 80.15 (\%) & 88.66 (\%) & \textbf{88.77} (\%) 
\end{tabular}}
\caption{Performance of different pre-training methods on the land cover classification task.}
\label{table:land_cover_classification}
\end{table}

As shown in Table~\ref{table:land_cover_classification}, our pre-training strategies lead to substantially higher performance than the ImageNet and CIFAR10 features. This demonstrates the robustness and wide range of applications our pre-training strategies possess.

\subsection{Experiments on Semantic Segmentation}
Previously, we explored image recognition in both pre-training and target tasks. In this section, we change the target task type to semantic segmentation to understand if image recognition pre-training can still boost the performance on semantic segmentation target task. Image recognition focuses on global features to associate an image $x_{i}$ with a label $w_{i}$. On the other hand, in semantic segmentation, local features play more important role in associating a pixel $x_{i}(m,n)$ with a label $w_{i}$ where $m$ and $n$ represent the column and row index. 
% Previously, \cite{mahajan2018exploring} explored the idea of pre-training on image recognition task and transferring the learned weights for the task of object detection. They report that such set up does not lead to significant performance increase as in the case where both pre-training and target tasks are image recognition tasks.

We first build a semantic segmentation network using the DenseNet121 model pre-trained on WikiSatNet. A typical segmentation model consists of encoder-decoder steps to generate segmentation maps with the same size to input images. In this case, we use the pre-trained weights $f_{v}$ as encoder and design a decoder architecture on top of $f_{v}$ as shown in Fig.~\ref{fig:segmentation_arch}. The proposed architecture is similar to U-Net architecture \cite{ronneberger2015u}, however, we use the DenseNet121 pre-trained on WikiSatNet in the decoder stage. This is important as it requires a complex network to learn from large-scale datasets such as WikiSatNet. On the other hand, the U-Net employs a shallow encoder, preventing us to pre-train it on the WikiSatNet. We perform experiments on the SpaceNet \cite{spacenet} semantic segmentation task on the satellite images. The SpaceNet contains training and test set from six different cities, and building and road masks for corresponding high resolution ($0.3$-$0.5$m GSD) DigitalGlobe images. In this study, we focus on the \emph{Rio} set and building masks. There are about 5000 training and 800 test images coming from the city of Rio de Janeiro. We experiment with varying number of training samples to quantify the learned representations in the case of using different amount of labeled samples in the target task. However, we keep the test set unchanged in our experiments. Table~\ref{table:semantic_segmentation_results} shows the Intersection-over-Union (IoU) scores of the proposed segmentation architecture (see Fig.~\ref{fig:segmentation_arch}) when pre-trained on ImageNet and WikiSatNet.

\begin{figure}[!h]
\centering
\includegraphics[width=0.45\textwidth]{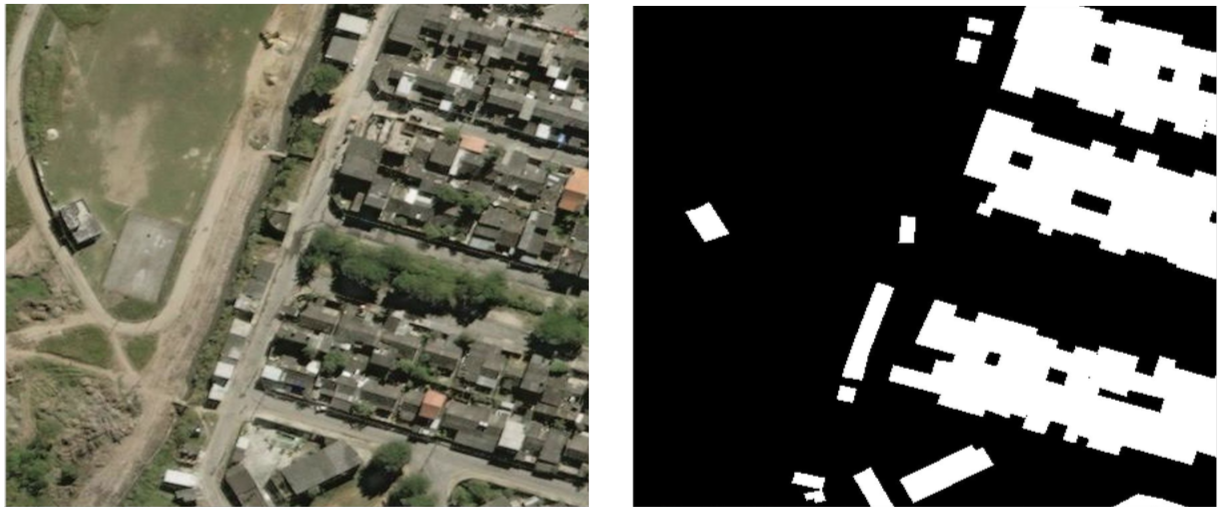}\\
\caption{An example of a building mask for a satellite image in SpaceNet \emph{Rio} dataset.}
\label{fig:building_footprints}
\end{figure}

\begin{table}[h]
\centering
\resizebox{0.48\textwidth}{!}{\begin{tabular}{@{}lccc@{}}
Model & \begin{tabular}[c]{@{}c@{}} From Scratch \end{tabular} & \begin{tabular}[c]{@{}c@{}} ImageNet \end{tabular} & \begin{tabular}[c]{@{}c@{}} WikiSatNet \\ \textit{Image2Text} \end{tabular} \\
\hline
\begin{tabular}[c]{@{}c@{}}200 Samples \end{tabular} & 42.11 (\%) & 50.75 (\%)  & \textbf{51.70} (\%) \\
\begin{tabular}[c]{@{}c@{}}500 Samples \end{tabular} & 48.98 (\%) & 54.63 (\%) & \textbf{55.41} (\%) \\
\begin{tabular}[c]{@{}c@{}}5000 Samples \end{tabular} & 57.21 (\%) & 59.63 (\%) & \textbf{59.74} (\%)
\end{tabular}}
\caption{The IoU scores of different pre-training methods on building segmentation task.}
\label{table:semantic_segmentation_results}
\end{table}

As shown in Table~\ref{table:semantic_segmentation_results}, the pre-training provides significant boost when fine-tuning on small amount of training samples (200, and 500 samples). However, pre-training on the WikiSatNet only achieves slightly higher IoU than ImageNet. These results are consistent with the previous studies where pre-training and target datasets contain different level tasks \cite{mahajan2018exploring}. For example, \cite{mahajan2018exploring} explored the idea of pre-training on image recognition task and transferring the learned weights for the task of object detection. They report that such set up does not lead to significant performance increase as in the case where both pre-training and target tasks are the same-level tasks (image recognition).

\section{Conclusion}
In this study, we proposed a novel combination of satellite images and crowdsourced annotations from geo-referenced Wikipedia articles. To the best of our knowledge, this is the first time that Wikipedia has been used this way. Our approach yields a large scale, multi-modal 
%\neal{be consistent with hyphens} 
dataset combining rich visual and textual information for millions of locations all over the world --- including additional languages beyond English will likely improve coverage even more. Leveraging paired multi-modal data, we proposed two different pre-training methods: (1) learning with weak labels, and (2) learning without weak labels using image to text matching. Both pre-training strategies lead to improved results on the recently released fMoW dataset consisting of large numbers of labeled samples. Our image to text matching model outperformed one pre-trained on ImageNet by $4.5\%$ when using around 350000 labeled samples; this increase in performance is substantially higher when there are fewer labeled samples available. 
%We believe this novel combination of satellite images and Wikipedia articles will open new research avenues both in computer science and the social sciences. 
%In this vein, fine-grained spatial information extracted from the articles could complement existing data sources, especially in the developing world where data is typically very scarce.

\bibliographystyle{plain}
\bibliography{references}
\end{document}